\documentclass[conference]{IEEEtran}
\IEEEoverridecommandlockouts

\usepackage{cite}
\usepackage{amsmath,amssymb,amsfonts}
\usepackage{algorithmic}
\usepackage{graphicx}
\usepackage{textcomp}
\usepackage{xcolor}
\usepackage{hyperref}

\usepackage{booktabs}       

\def\BibTeX{{\rm B\kern-.05em{\sc i\kern-.025em b}\kern-.08em
    T\kern-.1667em\lower.7ex\hbox{E}\kern-.125emX}}
\begin{document}

\title{ES-dRNN with Dynamic Attention for Short-Term Load Forecasting\\
}
\author{\IEEEauthorblockN{Slawek Smyl}
\IEEEauthorblockA{
\textit{Meta} \\
1 Hacker Way, Menlo Park, CA 94025, USA \\
slawek.smyl@gmail.com}
\and
\IEEEauthorblockN{Grzegorz Dudek}
    \IEEEauthorblockA{
    \textit{Department of Electrical Engineering} \\
	\textit{Czestochowa University of Technology}\\
	Czestochowa, Poland \\
	grzegorz.dudek@pcz.pl}
\and
\IEEEauthorblockN{Paweł Pełka}
    \IEEEauthorblockA{\textit{Department of Electrical Engineering} \\
	\textit{Czestochowa University of Technology}\\
	Czestochowa, Poland \\
	pawel.pelka@pcz.pl}	
}

\maketitle

\begin{abstract}
Short-term load forecasting (STLF) is a challenging problem due to the complex nature of the time series expressing multiple seasonality and varying variance. 
This paper proposes an extension of a hybrid forecasting model combining exponential smoothing and dilated recurrent neural network (ES-dRNN) with a mechanism for dynamic attention. We propose a new gated recurrent cell -- attentive dilated recurrent cell, which implements an attention mechanism for dynamic weighting of input vector components. The most relevant components are assigned greater weights, which are subsequently dynamically fine-tuned. This attention mechanism helps the model to select input information and, along with other mechanisms implemented in ES-dRNN, such as 
adaptive time series processing, cross-learning, and multiple dilation, leads to a significant improvement in accuracy when compared  to well-established statistical and state-of-the-art machine learning forecasting models. This was confirmed in the extensive experimental study concerning STLF for 35 European countries.
 
\end{abstract}


\section{Introduction}

STLF is a challenging problem due to the complex nature of the time series. Electricity load or demand time series exhibit three seasonal components (annual, weekly and daily), a nonlinear trend, varying variance and random fluctuations. The daily load pattern differs in shape depending on the day of the week and season of the year.
The load time series is strongly influenced by climatic, weather and economic conditions, all of which have a stochastic nature. All these properties place high demands on forecasting models. STLF is considered to be a very challenging problem, which is why it is often used for testing new models.

STLF is also very important in practice as an integral part of
power system control and scheduling. It is needed for the efficient and safe operation of power systems and supporting transactions of participants in deregulated electricity markets. Due to the importance and complexity of the STLF problem, 
a large number of different STLF models have been reported in the literature. They employ conventional statistical methods, computational intelligence and machine learning methods as well as hybrid solutions. The most popular new approaches for STLF are based on neural networks (NNs) \cite{Ben20}. They offer more possibilities than statistical models and exceed many of their limitations. These limitations include their linear  nature, limited ability to model complicated seasonal patterns, limited adaptability, a shortage  of expressive  power, and problems with capturing long-term dependencies and introducing exogenous variables.

Due to their flexibility and universal approximation property, NNs can model any nonlinear relationship and reflect process variability in an uncertain dynamic environment. Thus, they are often used for complex forecasting problems such as STLF. The classical NN architectures such as multilayer perceptron (MLP), radial basis function (RBF) NN, generalized regression neural network (GRNN), fuzzy counterpropagation NN, and self-organizing maps were compared as STLF models in \cite{Dud16a}. Multiple seasonality, which is a real problem for the forecasting models that usually requires decomposition or deseasonalisation, was solved in this study by appropriate representation of the time series defining patterns of the daily profiles. Thanks to this, the forecasting problem was simplified and 
it was possible to apply less complex neural models with a smaller number of parameters, which were more resistant to overfitting. Recently, classical MLP with pattern representation was replaced by randomized NN \cite{Dud21a}. Due to randomized learning, the training becomes much faster and easier and the numbers of parameters and hyperparameters to adjust is significantly reduced. At the same time, the accuracy of STLF improves in relation to MLP. Many STLF approaches combine neural models with effective optimization procedure and time series decomposition or a feature engineering method. For example, in \cite{Bas09}, the time series is decomposed into orthonormal series generated by a wavelet transform, and MLP is learned using a particle swarm optimization algorithm. In \cite{Hip10}, to control MLP complexity and to select input variables, a Bayesian approach was used. The Bayesian framework offered ways to avoid overfitting by regularisation, to decide on the number of neurons and to deal with the inputs by soft-pruning.

The rapid development of deep and recurrent NNs in recent years has led to the development of effective STLF methods. The new possibilities they offer, which are very attractive for forecasting models, include learning of representation, cross-learning on massive datasets and modeling temporary relationships in sequential data. Some examples of STLF models based on deep learning are: \cite{Che19}, where deep residual NNs were proposed and applied to probabilistic load forecasting using Monte Carlo dropout; \cite{Hos19}, where a multivariate fuzzy time series was converted into multi-channel images and processed by CNN to produce load forecasts; and \cite{Kon20}, where an improved deep belief network for STLF with demand-side management was proposed.

Recurrent NNs, which were designed for sequential data such as time series and text data are extremely useful for forecasting problems. Modern RNNs such as long-short term memory (LSTM) and gated recurrent unit (GRU) are distinguished by their ability to model both short and long-term dependencies in time series \cite{Hew21}.
Examples of RNN for STLF can be found in: \cite{Wan19}, where a model based on bi-directional LSTM and attention mechanism was proposed; \cite{Kon19}, where a STLF problem for individual residential households was addressed using LSTM; and \cite{Li21}, 
where the load is decomposed into different frequency components using the empirical-mode decomposition algorithm and then low-frequency components are predicted using linear regression while the high-frequency components are predicted using LSTM.

In our recent work \cite{Smy21}, motivated by new achievements in deep learning and RNNs \cite{Smy20}, we proposed a hybrid hierarchical STLF model combining exponential smoothing (ES) and dilated RNN (ES-dRNN). Its hybrid architecture improved ability to learn representation and explore hidden patterns. To deal with multiple seasonalities and temporal dependencies in time series, we proposed a new dilated recurrent cell (dRNNCell) and multiple dilated stacked RNN architecture. The simultaneous learning of both ES and dRNN components 
enables the entire model to be optimized while learning on multiple time series (cross-learning) enables ES-dRNN to capture the shared features of individual series. The model proposed in this work inherits ES-dRNN properties and extends it with a dynamic attention mechanism \cite{Vas17}. 
This mechanism allows the model to focus on relevant input information while the target functions for both point forecasts and predictive intervals are being modeled. 

The contribution of this study includes the following points:
\begin{enumerate}
    \item 
    We propose a new attentive dilated recurrent cell, adRNNCell, which implements an attention  mechanism for weighting the input information. It produces an internal attention vector which dynamically weights input vector components. This attention mechanism permits the recurrent cell to utilize the most relevant components of the input patterns in a flexible manner
    to improve the forecasting performance of the model. 

    \item
    We develop a hybrid exponential smoothing and dilated recurrent  NN model with attention based on adRNNCell. The model produces vectors of point forecasts and also predictive intervals. Due to its internal mechanisms such as dynamic attention, adaptive time series processing, cross-learning, and multiple dilations, the model can deal with complex time series expressing multiple seasonality and varying variance. 
    Our model is available as open-source code in the github repository \cite{rep22}.
    \item 
    We empirically demonstrate on real data sets of 35 European countries that our proposed model significantly outperforms in terms of STLF accuracy its predecessor \cite{Smy21} and 13 baseline forecasting models, including well-established statistical approaches and state-of-the-art machine learning approaches.

\end{enumerate} 

The rest of the work is organized as follows. In Section II, we present the architecture of the hybrid exponential  smoothing  and  dilated  recurrent  NN model with dynamic attention (ES-adRNN). Details of the proposed adRNNCell are described in Section III. The experimental framework used to evaluate the performance of the proposed approach is described in Section IV. Finally, Section V concludes the work.

\section{ES-adRNN Architecture}

\begin{figure}
	\centering
	\includegraphics[width=0.48\textwidth]{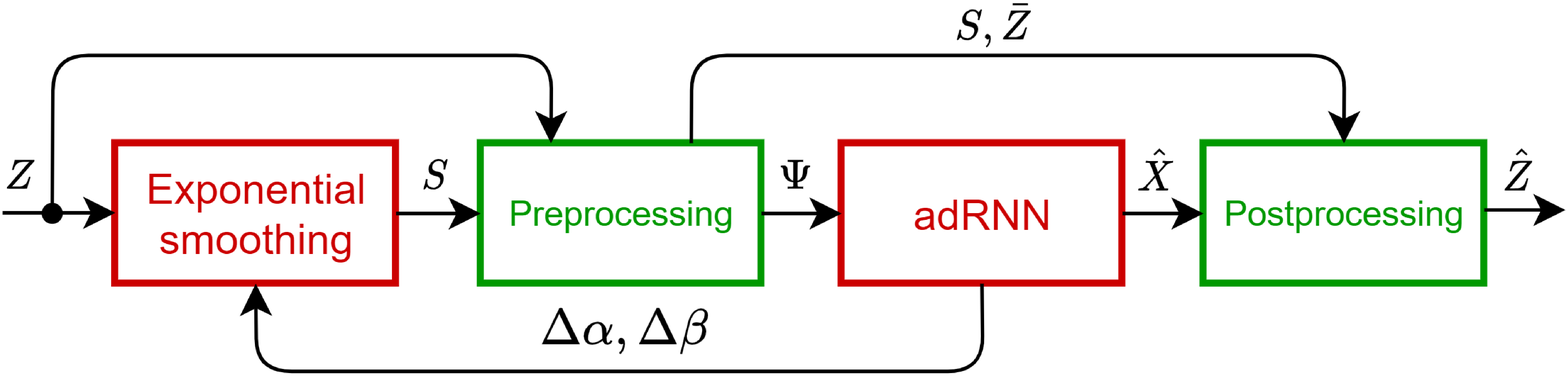}
	\caption{Block diagram of the proposed ES-adRNN forecasting model.} 
	\label{figdRNN}
\end{figure}

A hybrid exponential smoothing and dilated recurrent NN model with an attention mechanism, ES-adRNN, is composed of four main components as shown in Fig. \ref{figdRNN}. This architecture is similar to that of ES-dRNN proposed in \cite{Smy21}, except for the RNN component. In this study, we introduce RNN with a new gated recurrent cell, which is equipped with an attention mechanism. This new RNN component, adRNN, is described in details in Section \ref{adRNN}.  

The proposed model works in a recursive manner, generating daily pattern forecasts for the following days in the subsequent recursive steps $t$. The model input, $Z$, represents a set of $L$ time series: $\{\{z_\tau^l\}_{\tau=1}^{M_l}\}_{l=1}^L$, where $M_l$ is an $l$-th time series length. In our case, the series express hourly electricity demand for $L$ countries. The model learns simultaneously on $L$ time series, i.e. in a cross-learning mode \cite{Smy20}, which enables it to capture the shared features of the individual time series. 

ES decomposes each series into level and seasonal components. The seasonal components, $S$, are used by a preprocessing component to deseasonalize the original series. The series are also normalized and squashed to prepare the training set, $\Psi$, for adRNN. adRNN learns not only the main mapping function transforming inputs into electricity demands for the next day (preprocessed) but also their predictive intervals (PIs) and the corrections of ES smoothing parameters, $\Delta\alpha$ and $\Delta\beta$. A postprocessing component transforms forecasts of the preprocessed demand and PIs, $\hat{X}$, into real values, $\hat{Z}$. It uses for this seasonal components, $S$, and average values of input sequences, $\bar{Z}$, determined by the ES and preprocessing components.          

\subsection{Exponential Smoothing}

We use a simplified Holt-Winters multiplicative ES model, which decomposes the time series into two components: level component and seasonal component. A unique feature of the proposed approach is that the smoothing coefficients of the Holt-Winters model are not fixed, as in the typical case, but they are adapted in each recursive step $t$ as follows:

\begin{equation}
\begin{aligned}
\alpha_{t+1} = \sigma(I\alpha + \Delta\alpha_t)\\
\beta_{t+1} = \sigma(I\beta + \Delta\beta_t)
\label{eqab}
\end{aligned}
\end{equation} 
where $\alpha, \beta \in [0, 1]$ are smoothing coefficients, $I\alpha$ and $I\beta$ are initial values of the smoothing coefficients, 
$\Delta\alpha_t$ and $\Delta\beta_t$ are the corrections, 
and $\sigma$ denotes a sigmoid function.

The corrections are learned by adRNN simultaneously with the main mapping function. Thus, they have a dynamic character. They depend on the adRNN input representing the current time series characteristic and calendar variables. The dynamic version of the Holt-Winters multiplicative model takes the form:

\begin{equation}
\begin{aligned}
l_{t,\tau}=\alpha_t \frac{z_\tau}{s_{t,\tau}} + (1-\alpha_t)l_{t,\tau-1} \\
s_{t,\tau+168}=\beta_t \frac{z_\tau}{l_{t,\tau}} + (1-\beta_t)s_{t,\tau}
\label{eqls1}
\end{aligned}
\end{equation} 
where $\{z_\tau\}_{\tau=1}^M$ is a decomposed time series, $l_{t,\tau}$ denotes a level component and $s_{t,\tau}$ denotes a weekly seasonal component.

\subsection{Preprocessing and Postprocessing}

The preprocessing component prepares training patterns for adRNN. An input pattern represents a weekly period covered by moving window $\Delta^{in}$ of size 168 hours. An output pattern represents the forecasted daily sequence covered by moving window $\Delta^{out}$ of size 24 hours, which follows $\Delta^{in}$.
The windows are shifted by 24 hours to obtain subsequent input and output patterns:

\begin{equation}
\begin{aligned}
\textbf{x}_1^{in} &= [x_1, ..., x_{168}], &\textbf{x}_1^{out} = [x_{169}, ..., x_{192}], \\
 \textbf{x}_2^{in} &= [x_{25}, ..., x_{192}], &\textbf{x}_2^{out} = [x_{193}, ..., x_{216}], \\
 ...
\end{aligned}
\label{eqpt}
\end{equation}

The components of the $t$-th pair of patterns express deseasonalized, normalized and squashed time series sequences covered by the $t$-th pair of windows. They are determined as follows: 

\begin{equation}
x_\tau=\log{\frac{z_\tau}{\bar{z}_t \hat{s}_{t,\tau}} }
\label{eqxt}
\end{equation}
where $\tau \in \Delta^{in}_t \cup \Delta^{out}_t$, $\bar{z}_t$ is the average value in $\Delta^{in}_t$ and $\hat{s}_{t,\tau}$ is the seasonal component predicted by ES \eqref{eqls1} for recursive step $t$.

The $\log$ function in \eqref{eqxt} squashes the data to prevent outliers from adversely affecting model performance. Note that the seasonal component in \eqref{eqxt} is adapted in each recursive step $t$. This makes the training data dynamic and enables the model to learn data representation.     

Input patterns $\textbf{x}_t^{in}$ are the main component of the input vectors for adRNN learning. Additionally, to introduce more information related to the forecasted period, the input vectors include: (i) a seasonal vector 
predicted by ES for the $t$-th output period reduced by 1, i.e. $\hat{\textbf{s}}_t = [\hat{s}_{t,\tau}-1]_{\tau=24(t-1)+169}^{24(t-1)+192}$,
(ii) a current level of the time series, $\log_{10}(\bar{z}_t)$, and (iii) calendar variables, 
$\textbf{d}_t^{w} \in \{0, 1\}^7, \textbf{d}_t^{m} \in \{0, 1\}^{31}$ and $\textbf{d}_t^{y} \in \{0, 1\}^{52}$, as binary one-hot vectors encoding day of the week, day of the month and week of the year for the forecasted day, respectively. The input vector takes the form: 

\begin{equation}
\textbf{x}_t^{in'}= [\textbf{x}_t^{in},\, \hat{\textbf{s}}_t,\, \log_{10}(\bar{z}_t),\, \textbf{d}_t^{w},\, \textbf{d}_t^{m},\, \textbf{d}_t^{y}] 
\label{eqxp}
\end{equation} 

The daily patterns forecasted by adRNN, $\hat{\textbf{x}}_t^{out} = [\hat{x}_\tau]_{\tau \in \Delta^{out}_t}$, are transformed by the postprocessing component to obtain real forecasts of the hourly electricity demand. For this, transformed equation \eqref{eqxt} is used:

\begin{equation}
\hat{z}_\tau=\exp (\hat{x}_\tau) \bar{z}_t\hat{s}_{t,\tau}
\label{eqzt}
\end{equation} 
where $\tau \in \Delta^{out}_t$.

\subsection{Loss Function}

The proposed model predicts hourly electricity demands, $z_\tau$, and their PIs in the form of two quantiles of orders $\underline{q}$ (lower) and $\overline{q}$ (upper). To optimize both point forecasts and PIs, we use a loss function in the form \cite{Smy21}:

\begin{equation}
L_\tau =
\rho(z'_\tau, \hat{z}'_{q^*,\tau}) + \gamma [
\rho (z'_\tau, \hat{{z}}'_{\underline{q},\tau}) + 
\rho (z'_\tau, \hat{{z}}'_{\overline{q},\tau})]
\label{eqlss}
\end{equation}
where $\rho$ denotes a pinball loss function:

\begin{equation}
\rho(z, \hat{z}_q) =
\begin{cases}
(z-\hat{z}_q)q       & \text{if } z \geq \hat{z}_q\\
(z-\hat{z}_q)(q-1)  &\text{if } z < \hat{z}_q 
\end{cases}
\label{eqrho}
\end{equation}
$z$ is an actual value; $\hat{z}_q$ is a forecasted value of $q$-th quantile; $q \in (0, 1)$ is a quantile order; $q^*=0.5$ corresponds to the median; 
$z'_\tau = z_\tau / \bar{z}_t$ is a normalized actual time series value from the output window $\Delta^{out}_t$;
$\hat{z}'_{q^*,\tau} = \exp(\hat{x}_\tau)\hat{s}_{t,\tau}$ is a forecasted value of $z'_\tau$; 
$\underline{q}$ and $\overline{q}$ are the quantile orders for the lower and upper bounds of PI, respectively; 
$\hat{{z}}'_{\underline{q},\tau} = \exp(\hat{\underline{x}}_\tau)\hat{s}_{t,\tau}$ is a forecasted value of $\underline{q}$-quantile of $z'_\tau$;
$\hat{{z}}'_{\overline{q},\tau} = \exp(\hat{\bar{x}}_\tau)\hat{s}_{t,\tau}$ is a forecasted value of $\overline{q}$-quantile of $z'_\tau$; and $\gamma \geq 0$ is a controlling parameter.

Loss function \eqref{eqlss} uses normalized values of the time series, $z'_\tau$, to bring the errors calculated for different time series to the same level, as this is important in cross-learning. 

The first component in \eqref{eqlss}, $\rho(z'_\tau, \hat{z}'_{q^*,\tau})$, 
is almost (see below) a symmetrical loss for the point forecast (normalized) while the second and third components, $\rho (z'_\tau, \hat{{z}}'_{\underline{q},\tau})$ and $\rho (z'_\tau, \hat{{z}}'_{\overline{q},\tau})$, are asymmetrical losses for the quantiles. The asymmetry level, which determines PI, results from the quantile orders. 
Parameter $\gamma$ enables us to control the impact of PI-related components on the loss value. Its typical value ranges from 0.1 to 0.5. 

It is worth noting that the pinball loss gives us the opportunity to reduce the forecast bias by penalizing positive and negative deviations differently. When the model tends to have a positive or negative bias, we can reduce the bias by introducing $q^*$ smaller or larger than $0.5$, respectively (see \cite{Smy20, Dud21}).

\section{Dilated RNN with Dynamic Attention}
\label{adRNN}

\begin{figure}
	\centering
	\includegraphics[width=0.35\textwidth]{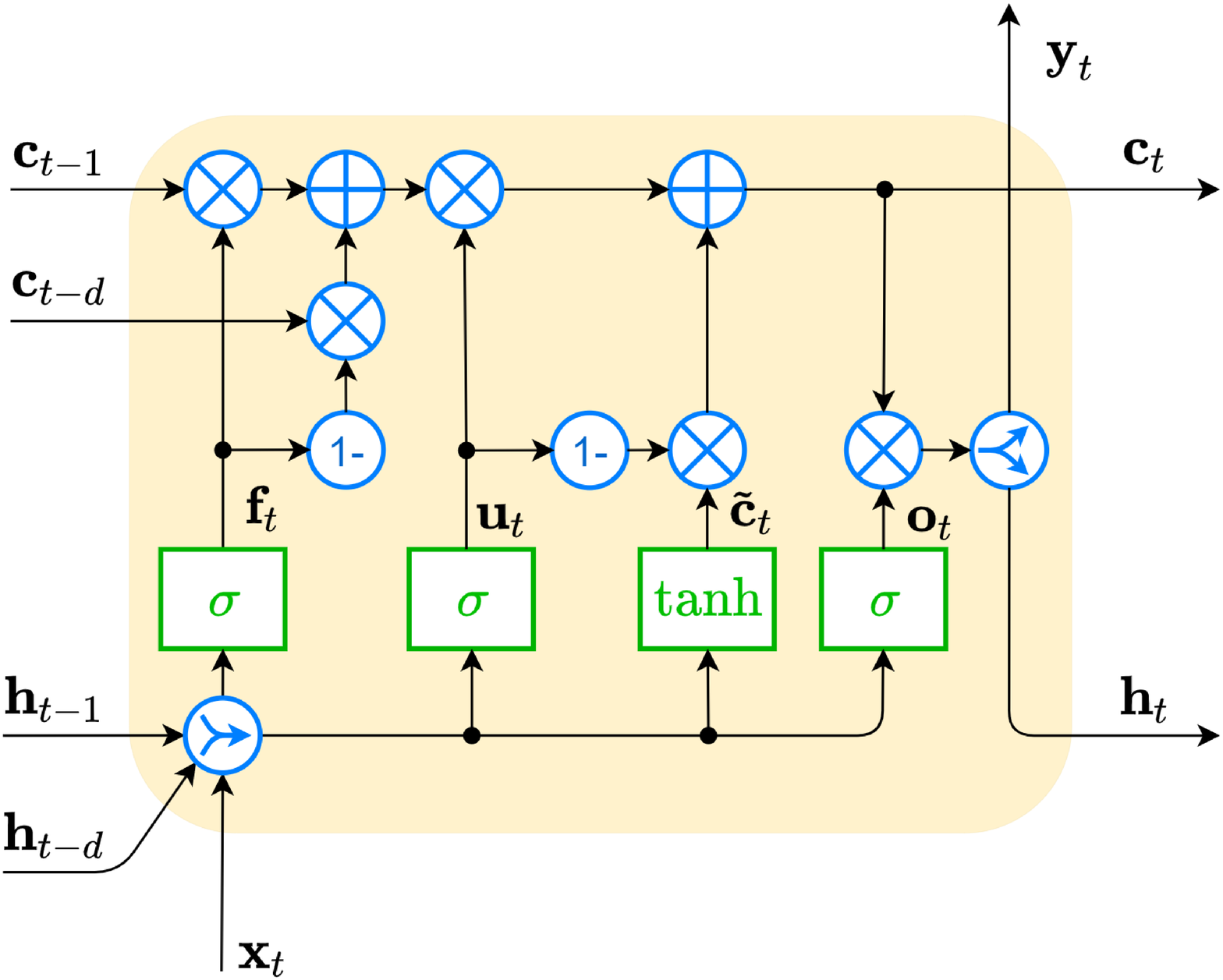}
    \includegraphics[width=0.35\textwidth]{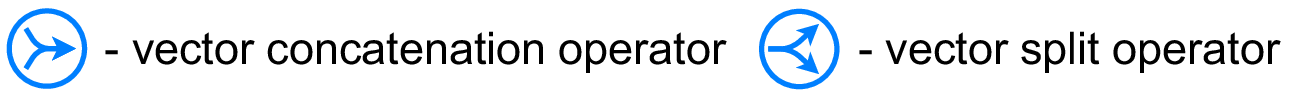}
    \caption{Dilated recurrent cell dRNNCell.} 
    \label{figCell1}
\end{figure}

The RNN component of ES-adRNN employs a dilated RNN cell (dRNNCell) introduced in \cite{Smy21}. This cell is shown in Fig. \ref{figCell1}. It was designed for STLF to deal with a complex seasonal pattern expressing three seasonalities. The distinguishing features of dRNNCell are:
\begin{itemize}
    \item dRNNCell is fed by two cell states ($c$-states) and two controlling states ($h$-states). They represent recent states, $\textbf{c}_{t-1}$, $\textbf{h}_{t-1}$, and delayed states, $\textbf{c}_{t-d}$, $\textbf{h}_{t-d}$, $d>1$.  
    \item The output of dRNNCell is split into "real output" $\textbf{y}_t$, which goes to the next layer, and a controlling output $\textbf{h}_t$, which is an input to the gating mechanism in the following time steps. 
\end{itemize}
   
\begin{figure}
	\centering
    \includegraphics[width=0.35\textwidth]{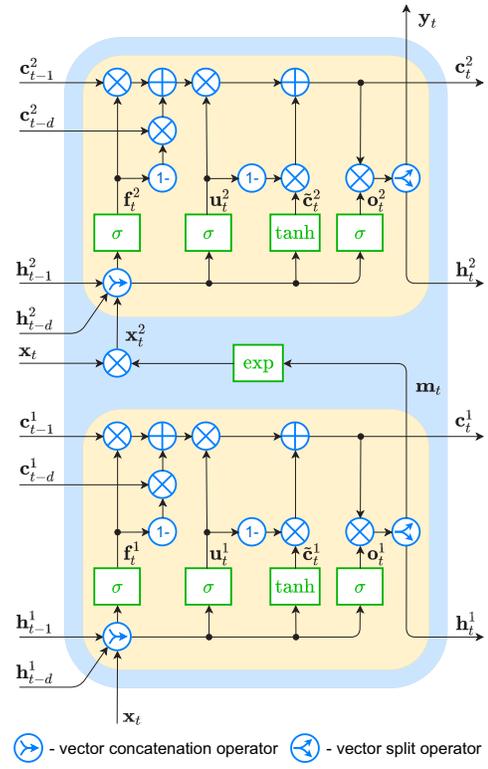}
    \includegraphics[width=0.35\textwidth]{Cell_legend.eps}
    \caption{Attentive dilated recurrent cell adRNNCell.} 
    \label{figCell2}
\end{figure}

In this study, we combine two dRNNCells to obtain a more efficient gated recurrent cell, which is able to preprocess dynamically the input data. A new cell, attentive dilated RNN cell (adRNNCell), introduces an attention mechanism for weighting the input information. It is shown in Fig. \ref{figCell2}. The first cell (at the bottom of the figure) produces attention vector $\textbf{m}_t$ of the same length as input vector $\textbf{x}_t$. The components of $\textbf{m}_t$, after processing by $\exp$ function, are treated as weights for the inputs collected in vector $\textbf{x}_t$. Thus, they strengthen or weaken the particular inputs to the second cell (at the top of the figure). Note that $\textbf{m}_t$ has a dynamical character - the weights are adjusted to the current inputs at time $t$. Both cells learn simultaneously. Based on preprocessed input vector, $\textbf{x}_t^2$, the second cell produces vector $\textbf{y}_t$, which feeds the next layer.  

dRNNCell as a separate cell and as a component cell of adRNNCell, uses three gates. Namely, a fusion gate $f$, update gate $u$, and output gate $o$. The gates transform the input vectors, $\textbf{x}_t$, $\textbf{h}_{t-1}$ and $\textbf{h}_{t-d}$, using sigmoid nonlinearity $\sigma$. A candidate $c$-state, $\tilde{\textbf{c}}_t$, is produced by transforming input vectors using $\tanh$ nonlinearity. 
The dRNNCell formulation is as follows:

\begin{equation}
\textbf{f}_t= \sigma(\textbf{W}_f \textbf{x}_t+\textbf{V}_f\textbf{h}_{t-1}+\textbf{U}_f\textbf{h}_{t-d}+ \textbf{b}_f)
\label{eqp}
\end{equation} 
\begin{equation}
\textbf{u}_t= \sigma(\textbf{W}_u \textbf{x}_t+\textbf{V}_u\textbf{h}_{t-1}+\textbf{U}_u\textbf{h}_{t-d}+ \textbf{b}_u)
\label{eqf}
\end{equation} 
\begin{equation}
\textbf{o}_t= \sigma(\textbf{W}_o \textbf{x}_t+\textbf{V}_o\textbf{h}_{t-1}+\textbf{U}_o\textbf{h}_{t-d}+ \textbf{b}_o)
\label{eqo}
\end{equation} 
\begin{equation}
\tilde{\textbf{c}}_t= \tanh(\textbf{W}_c \textbf{x}_t+\textbf{V}_c\textbf{h}_{t-1}+\textbf{U}_c\textbf{h}_{t-d}+ \textbf{b}_c)
\label{eqg}
\end{equation} 
where subscript $t$ denotes a time step; $\sigma$ denotes a logistic sigmoid function; $\textbf{x}_t$ is an input vector; $\textbf{h}_{t-1}$ and $\textbf{h}_{t-d}$ are recent and delayed controlling states; $d>1$ is a dilation; $\textbf{W}$, $\textbf{V}$ and $\textbf{U}$ are weight matrices; and $\textbf{b}$ are bias vectors.

The $c$-state is calculated from the recent ($\textbf{c}_{t-1}$), delayed ($\textbf{c}_{t-d}$) and candidate ($\tilde{\textbf{c}}_t$) states as follows:   

\begin{equation}
\textbf{c}_t = \textbf{u}_t  \otimes  
\left(\textbf{f}_t \otimes \textbf{c}_{t-1}  +
\left(1-\textbf{f}_t\right) \otimes \textbf{c}_{t-d}\right) +
(1-\textbf{u}_t) \otimes \tilde{\textbf{c}}_t
\label{eqc}
\end{equation}
where $\otimes$ denotes the Hadamard product (element-wise product).

Note that the $c$-state is a weighted combination of past $c$-states and a new candidate state. Fusion vector $\textbf{f}_t$ decides in what proportion the recent and delayed $c$-states are mixed, while update vector $\textbf{u}_t$ 
decides about the share of old and new information in the resulting $c$-state.

Based on state $\textbf{c}_t$, the output vectors $\textbf{h}_t$ and $\textbf{y}_t$ (or $\textbf{m}_t$) of the cells shown in Fig. \ref{figCell2} are determined as follows:

\begin{equation}
\textbf{h}_t^{1\prime} = \textbf{o}_t^1   \otimes  \textbf{c}_t^1,  \textbf{m}_t= [h_{t,1}^{1\prime}, ..., h_{t,s_m}^{1\prime}], 
\textbf{h}_t^1= [h_{t,s_m+1}^{1\prime}, ..., h_{t,s_m+s_h}^{1\prime}] 
\label{eqh1}
\end{equation}
\begin{equation}
\textbf{h}_t^{2\prime} = \textbf{o}_t^2   \otimes  \textbf{c}_t^2, \textbf{y}_t= [h_{t,1}^{2\prime}, ..., h_{t,s_y}^{2\prime}],
\textbf{h}_t^2= [h_{t,s_y+1}^{2\prime}, ..., h_{t,s_y+s_q}^{2\prime}] 
\label{eqh2}
\end{equation}
where $s_y, s_h, s_m$ and $s_q$ denote the lengths of vectors $\textbf{y}_t, \textbf{h}_t^1, \textbf{m}_t$ and $\textbf{h}_t^2$, respectively (in our implementation $s_h$ and $s_q$ are the same).

\begin{figure}
	\centering
    \includegraphics[width=0.48\textwidth]{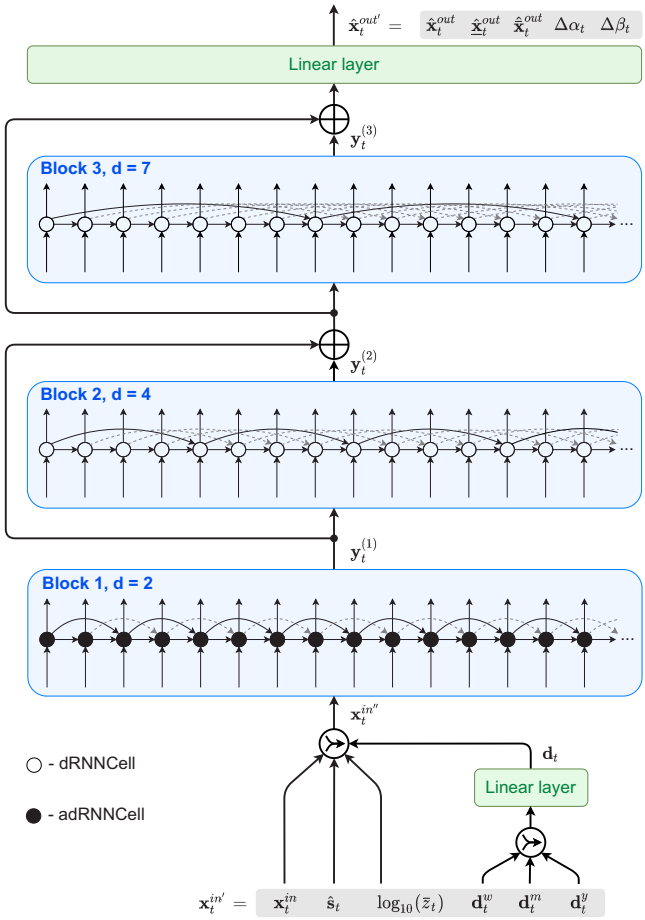}
    \caption{adRNN architecture.} 
    \label{figRNN}
\end{figure}

Fig. \ref{figRNN} shows the adRNN architecture. It is composed of three single-layer blocks. Block 1 consists of adRNNCell dilated 2, while blocks 2 and 3 consist of dRNNCells dilated 4 and 7, respectively. Note that the cells are fed by delayed states of different dilations in different layers. Delayed connections enable the direct input to the cell of information from a few time steps ago. This can be useful in modeling seasonal dependencies. In our architecture, we use multiple dilated recurrent layers stacked with hierarchical dilations to model the temporal dependencies of different scales. To facilitate the learning process, we use ResNet-style shortcuts between blocks \cite{He16}.

The inputs representing the calendar variables, $\textbf{d}_t^{w}$, $\textbf{d}_t^{m}$ and $\textbf{d}_t^{y}$, are embedded using a linear layer into $d$-dimensional continuous vector $\textbf{d}_t$. This reduces input dimensionality and meaningfully represents sparse binary vectors in the embedding space. The embedding is learned along with the model itself.

The output layer projects linearly vector $\textbf{y}_t^{(3)}$ produced by the third block to output vector $\hat{\textbf{x}}_t^{out'}$ composed as follows:

\begin{equation}
\hat{\textbf{x}}_t^{out'}= [\hat{\textbf{x}}_t^{out},\, 
\hat{\underline{\textbf{x}}}_t^{out},\,
\hat{\bar{\textbf{x}}}_t^{out},\,
\Delta{\alpha_t},\,
\Delta{\beta_t}]
\label{eqout}
\end{equation} 
where $\hat{\textbf{x}}_t^{out} = [\hat{x}_\tau]_{\tau \in \Delta^{out}_t}$ is a forecasted output pattern; $\hat{\underline{\textbf{x}}}_t^{out} = [\hat{\underline{x}}_\tau]_{\tau \in \Delta^{out}_t}$ is a vector of lower bounds of PI; $\hat{\bar{\textbf{x}}}_t^{out} = [\hat{\bar{x}}_\tau]_{\tau \in \Delta^{out}_t}$ is a vector of upper bounds of PI; $\Delta{\alpha_t}$ and $\Delta{\beta_t}$ are corrections for smoothing coefficients.

\section{Experimental Study}

In this section, to evaluate ES-adRNN forecast accuracy, we consider STLF for 35 European countries. The data set includes real-world hourly electrical load time series from the period 2016-2018. The data was collected from ENTSO-E repository, www.entsoe.eu/data/power-stats/ (we share this data with the ES-adRNN code in our github repository \cite{rep22}). The data provides a variety of time series with different properties such as different levels, trends, variance and daily shapes (see Section II in \cite{Smy21} where these time series are analysed). This great variety of time series makes the results more reliable.

We consider a one day-ahead forecasting problem: prediction of the load profile (24 hourly values) for each day of 2018 based on historical data. This was performed for each country with the exception of three countries for which data for the last month (Estonia and Italy) or the last two months (Latvia) of 2018 was unavailable. For these three countries, the test periods were shorter.
The ES-adRNN model was optimized on data from the period 2016-17. 
We perform STLF in two variants: using ES-adRNN as an individual model and using an ensemble of five  ES-adRNN base models. We denote the latter variant by ES-adRNNe.
As performance metrics we use: mean absolute percentage error (MAPE), median of APE (MdAPE), interquartile range of APE (IqrAPE), root mean square error (RMSE), mean PE (MPE), and standard deviation of PE (StdPE).


\subsection{Training and Optimization Setup}

Our learning process, schedule of hyper-parameter changes is organized around a notion of an epoch. This is usually defined as using all the training data once. Our definition here is based on the number of updates or processed batches, as during training we step $l_o$ times on a batch (with random assignment of series) and for each batch execute a single update based on the average error. Our aim is to define the epoch as the number of updates which brings in a meaningful change in the learning process, (we use 2500), and because the data set contains a small number of series, a single epoch is actually composed with $n_o$ number "sub-epochs", defined in the traditional fashion as one scan of all available data. An additional factor is the batch size: when it grows, the number of updates per sub-epoch diminishes, so the number of the sub-epochs needs to grow. However, in our experience the linear growth is too fast, and we use a sub-linear formula.

During each epoch a number of updates is executed, guided by the average error accumulated by executing $l_o$ (e.g. 50) forward steps, moving by one day, on a batch. The starting point is chosen randomly; the batches include random $b$ series. The model is trained using Adam optimizer.

The $d$-dilated recurrent cells operate as described above only after $d$ steps, because only after $d$ steps are the delayed states available. Additionally, the Holt-Winters formulas require at least twice the seasonality steps to stabilize, so the system uses several weeks ($w_o$) at the beginning of each batch as a warm-up period, during which all the ES and RNN calculations take place, with the exception of the training errors, which are not calculated.
Similarly, an even longer warm-up period $w_s$ is applied when producing the test results.

We use a schedule of increasing batch sizes and decreasing learning rates proposed in \cite{Smi18}. We start with a small batch size of 2, and increase it, although only once, due to the small number of series, to 5 at epoch 4. 
We use another schedule of decreasing learning rates, which has a similar, but not exactly  the same, effect as increasing the batch size: it allows the validation error to be  further reduced. We use the following schedule: $3\cdot10^{-3}$ (epochs 1-4), $10^{-3}$ (epoch 5), $3\cdot10^{-4}$ (epoch 6), $10^{-4}$ (epochs 7-9).

The sizes of $c$-state and $h$-state were  $s_c=100$, $s_h=s_q=40$. These values were obtained by experimentation starting with $s_c=50$, $s_h=20$ and doubling it 3 times.

Finally, as described in Section II C, the pinball loss function was utilized, with three different quantile values $q$,  to achieve quantile regression for 0.5, 0.05, and 0.95. The actual values for $q^*$, $\underline{q}$, and $\overline{q}$ were slightly different: 0.485, 0.035, 0.96. These values were arrived at by experimentation, reducing the bias of the center value, and fine-tuning the percentage of exceedance for PIs.

Other hyperparameters were selected as described in \cite{Smy21}.

\subsection{ Baseline Models}

As baseline models we employ:
\begin{itemize}
\item Naive -- naive model in the form: the forecasted demand profile for day $i$ is the same as the profile for day $i-7$
\item ARIMA -- autoregressive integrated moving average model \cite{Dud15},
\item ES -- exponential smoothing model \cite{Dud15},
\item Prophet -- modular additive regression model with nonlinear trend and seasonal components \cite{Tay18},
\item FNM -- fuzzy neighborhood model \cite{Dud15}
\item GRNN -- general regression NN \cite{Dud16a},
\item MLP -- perceptron with a single hidden layer and sigmoid nonlinearities \cite{Dud16a},
\item SVM -- linear epsilon insensitive support vector machine ($\epsilon$-SVM) \cite{Pel21},
\item LSTM -- long short-term memory \cite{Pel20},
\item ANFIS -- adaptive neuro-fuzzy inference system \cite{Pel18},
\item MTGNN -- graph NN for multivariate time series forecasting \cite{Wu20},
\item ES-dRNN -- hybrid exponential smoothing and dilated recurrent NN model \cite{Smy21},
\item ES-dRNNe -- ensemble of five ES-dRNN base models \cite{Smy21}.
\end{itemize}

The baseline models include classical statistical models (ARIMA, ES), new statistical models (Prophet), nonparametric pattern-based machine learning models (FNM, N-WE), classical machine learning models (GRNN, MLP, SVM, ANFIS) and new recurrent and deep NN architectures (LSTM, MTGNN). They also  include the predecessor of the proposed model, i.e. ES-dRNN and its ensemble variant ES-dRNNe. 

\subsection{Results}

\begin{figure}
	\centering
	\includegraphics[width=0.4\textwidth]{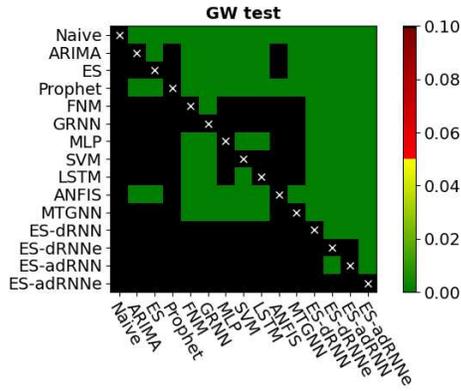}
	\caption{Results of the Giacomini-White test.} 
	\label{GW}
\end{figure}

Table \ref{tabEr} summarizes the forecasting quality metrics averaged over the 35 countries. It is clear from this table that ES-adRNNe outperforms all other models in terms of accuracy. It shows the lowest MAPE, MdAPE and RMSE. To confirm the best performance of ES-adRNNe, we perform a pairwise one-sided Giacomini-White test (GM test) for conditional predictive ability \cite{Gia06} (we used the multivariate variant of the GW test implemented in https://github.com/jeslago/epftoolbox \cite{Lag21}). Fig. \ref{GW} demonstrates results of this test as a heat map representing the obtained $p$-values. The closer the $p$-values are to zero
the significantly more accurate the forecasts produced by the model on the $X$-axis are than the forecasts produced by the model on the $Y$-axis. The black color is for $p$-values larger than 0.10, indicating rejection of the hypothesis that the model on the $X$-axis is more accurate than the model on the $Y$-axis. Fig. \ref{GW} clearly show the state-of-the-art performance of ES-adRNNe and ES-adRNN. ES-adRNNe performed significantly better in terms of accuracy than all the other comparative models. ES-adRNN is second only to the ES-dRNNe.

\begin{table}[]
	\caption{Forecasting quality metrics.}
	\begin{tabular}{lcccrrc}
		\toprule
		& MAPE  & MdAPE & IqrAPE & RMSE  & MPE   & StdPE \\
		\midrule    
    Naive & 5.08  & 4.84  & 3.32  & 704.34 & -0.26 & 7.91 \\
    ARIMA & 3.30  & 3.01  & 3.00  & 475.09 & \textbf{-0.01} & 5.31 \\
    ES    & 3.11  & 2.88  & 2.73  & 439.26 & \textbf{0.01} & 5.13 \\
    Prophet & 4.53  & 4.32  & 3.03  & 619.39 & -0.13 & 6.82 \\
    FNM   & 2.50  & 2.30  & 2.29  & 334.08 & -0.11 & 4.27 \\
    GRNN  & 2.48  & 2.28  & 2.27  & 332.91 & -0.11 & 4.25 \\
    MLP   & 3.05  & 2.78  & 2.94  & 419.01 & -0.04 & 5.07 \\
    SVM   & 2.55  & 2.29  & 2.52  & 357.24 & -0.13 & 4.37 \\
    LSTM  & 2.76  & 2.57  & 2.52  & 381.76 & 0.02  & 4.47 \\
    ANFIS & 3.65  & 3.17  & 3.66  & 507.08 & -0.10 & 6.43 \\
    MTGNN & 2.99  & 2.74  & 2.69  & 405.18 & -0.47 & 4.85 \\
    ES-dRNN & 2.33  & 2.13  & 2.23  & 319.04 & -0.20 & 3.90 \\
    ES-dRNNe & 2.25  & 2.05  & 2.17  & 309.88 & -0.20 & 3.79 \\
    ES-adRNN & 2.28  & 2.08  & 2.19  & 315.44 & -0.16 & 3.82 \\
    ES-adRNNe & \textbf{2.20} & \textbf{2.01} & \textbf{2.13} & \textbf{303.70} & -0.13 & \textbf{3.71} \\
		\bottomrule
	\end{tabular}
	\label{tabEr}
\end{table}

IqrAPE and StdPE shown in Table \ref{tabEr} measure the dispersion of APE and PE, respectively. Note that ES-adRNNe produced the least dispersed forecasts. The individual version, ES-adRNN, is close behind it.
MPE is a measure of the forecast bias. Note that most of the models including ES-adRNN and ES-adRNNe produced negatively biased forecasts, which means overprediction. The proposed model is equipped with a mechanism for controlling the bias included in the loss function \eqref{eqlss}, so the bias can be reduced. But reduction in bias can lead to a decrease in forecast accuracy due to overfitting, so we purposely avoided it. 

\begin{figure}
	\centering
	\includegraphics[width=0.237\textwidth]{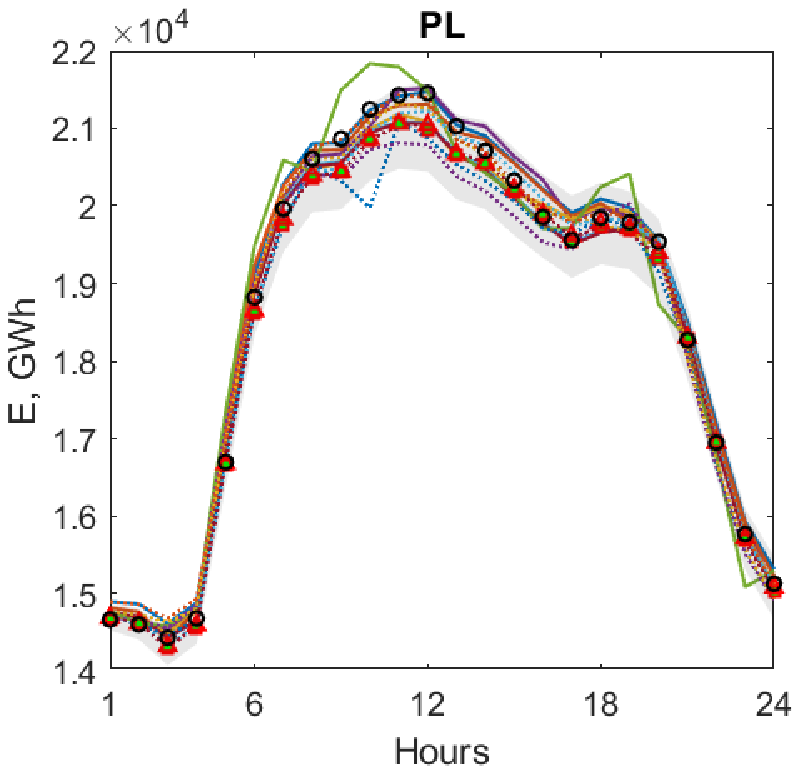}
	\includegraphics[width=0.237\textwidth]{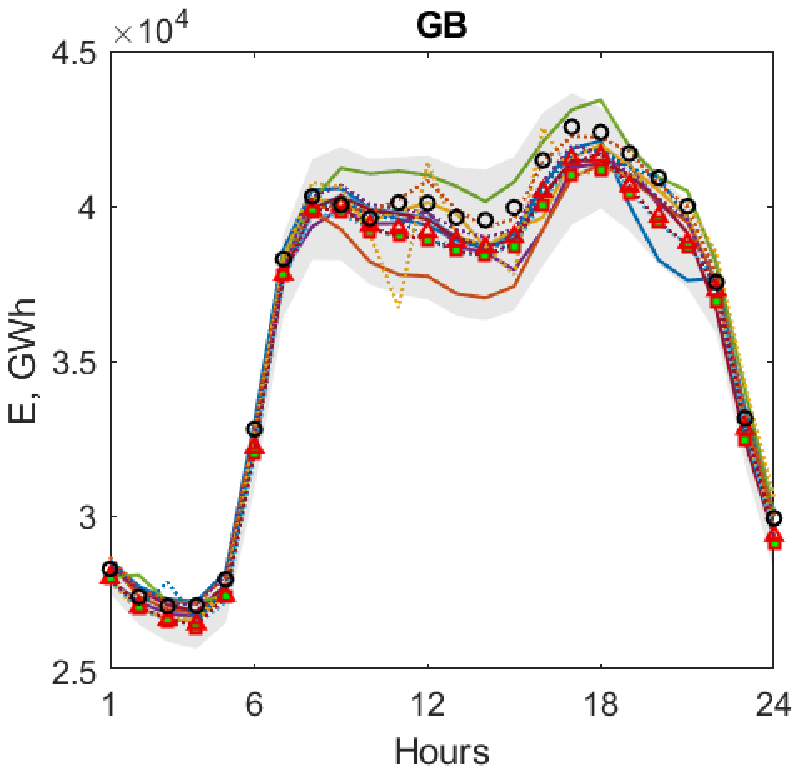}
	\includegraphics[width=0.237\textwidth]{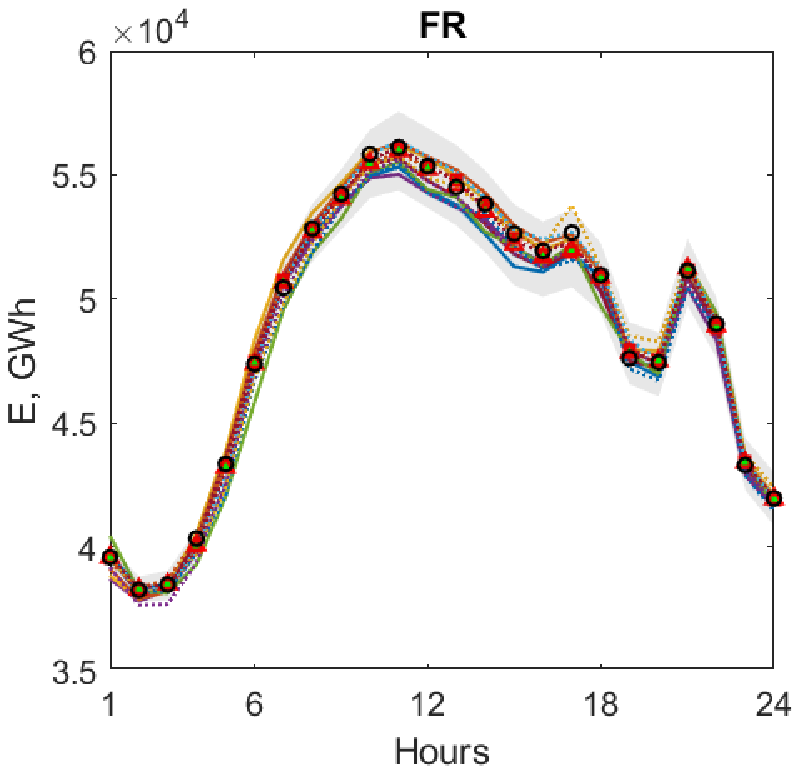}
	\includegraphics[width=0.237\textwidth]{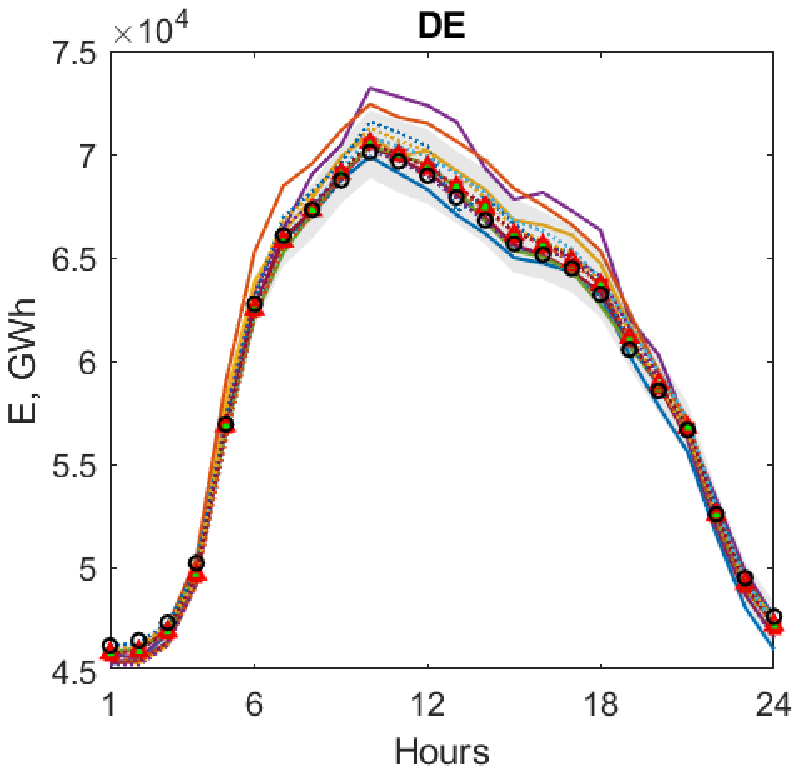}
	\includegraphics[width=0.237\textwidth]{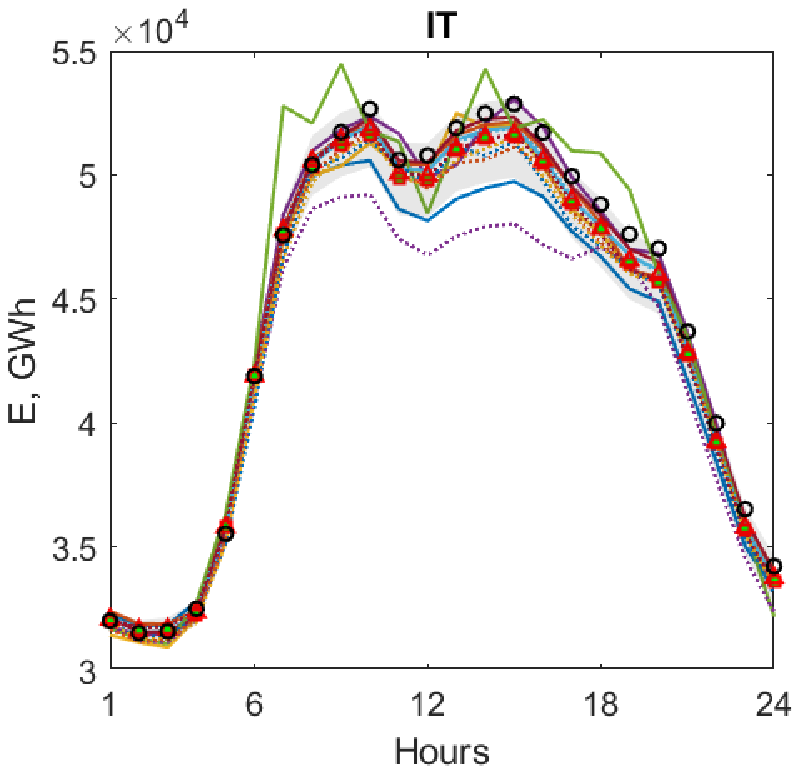}
	\includegraphics[width=0.237\textwidth]{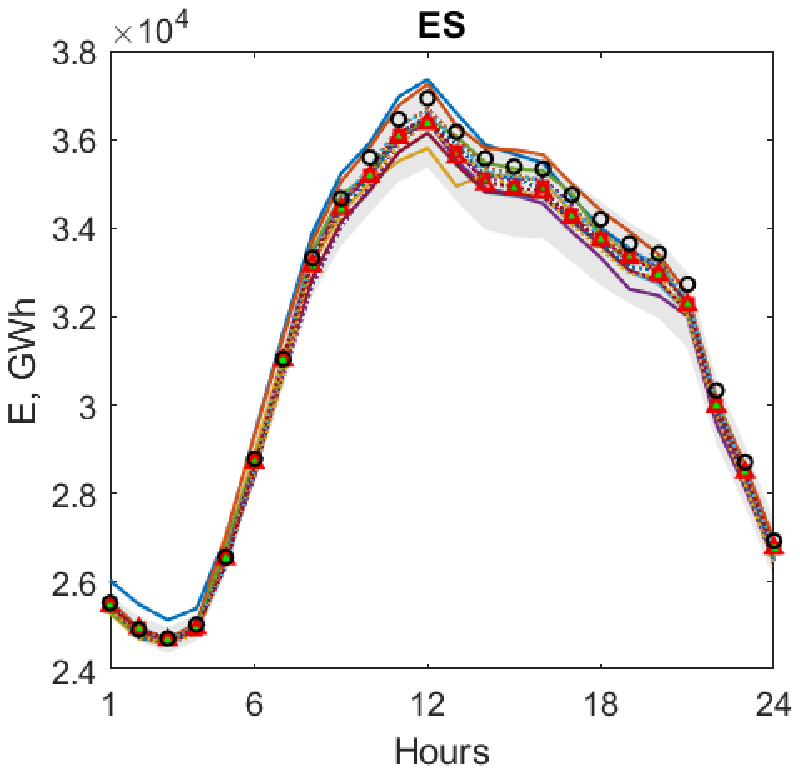}
	\includegraphics[width=0.48\textwidth]{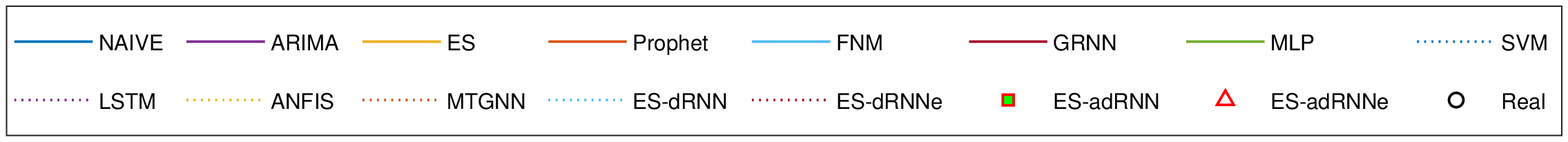}
	\caption{Examples of the forecasted daily profiles. 90\% PIs for ES-adRNNe are shown as gray-shaded areas.} 
	\label{figPro}
\end{figure}

Fig. \ref{figPro} shows several examples of forecasted daily patterns. PIs are also shown for ES-adRNNe. To evaluate the accuracy of the PIs predicted by this model, we calculate the number of forecasts lying inside, above and below of their PIs. We achieved: $90.18\% \pm 2.86\%$ forecasts are inside PIs, $4.74\% \pm 1.47\%$ are below PIs and $5.08\% \pm 1.61\%$ are above PIs. These values correspond to the assumed PIs of 90\% with lower and upper bounds $\underline{q}=0.05$ and $\overline{q}=0.95$, respectively.

\section{Conclusion}
STLF is challenging due to multiple seasonality, nonlinear trend and variable variance. To deal with this problem our model combines ES for time series preprocessing and RNN for capturing both short and long-term dependencies in time series. It is equipped with many useful mechanisms and procedures such as cross-learning on many time series, 
common learning procedure for ES and RNN,
adjusting of ES parameters by RNN,
on-the-fly deseasonalization,
dilated  recurrent  cells,
ResNet-style shortcuts, and
extended input vectors. 

In this work, we extend further the model by adding dynamic attention. We introduce a new type of the gated recurrent cell, adRNNCell, which implements an attention  mechanism for weighting the input information. This mechanism permits the cell to utilize the most relevant components of the input patterns in a flexible manner to improve the forecasting performance of the model.  

As the experimental study showed, dynamic attention significantly improves the accuracy of forecasting. The proposed ES-adRNNe outperformed statistical and machine learning models including its predecessor ES-dRNNe.
Note that ensembling in our approach does not require additional effort related to the selection of additional hyperparameters such as a parameter for controlling learners' diversity. The diversity is provided by the random initialization of the models. The
great advantages of ES-adRNNe are its ability to deal with raw time series, without requiring their decomposition, and its ability to produce both point forecasts and PIs. PIs with a specified probability coverage give very valuable information about the uncertainty of the prediction.

In further research, we plan to enrich the input information with a learned context vector. This represents information extracted from other time series, which can help predict a given time series.

\end{document}